\documentclass[sn-mathphys]{sn-jnl}

\usepackage[shortlabels]{enumitem}
\usepackage{amsmath}
\usepackage{amssymb}
\usepackage[T1]{fontenc}
\usepackage{array}
\usepackage{siunitx}

\newcolumntype{C}[1]{>{\centering\let\newline\\\arraybackslash\hspace{0pt}}m{#1}}
\jyear{2021}%

\theoremstyle{thmstyleone}%
\theoremstyle{thmstyletwo}%

\theoremstyle{thmstylethree}%

\begin{document}

\title[Optimizing Data Augmentation Policy Through Random Unidimensional Search]{Optimizing Data Augmentation Policy Through Random Unidimensional Search}


\author*[1,2]{\fnm{Xiaomeng} \sur{Dong}}\email{Xiaomeng.Dong@ge.com}

\author[1]{\fnm{Michael} \sur{Potter}}

\author[1]{\fnm{Gaurav} \sur{Kumar}}

\author[1]{\fnm{Yun-Chan} \sur{Tsai}}

\author[1]{\fnm{V. Ratna} \sur{Saripalli}}

\author[2]{\fnm{Theodore} \sur{Trafalis}}

\affil[1]{\orgname{GE HealthCare},\country{USA}}

\affil[2]{\orgname{University of Oklahoma}, \country{USA}}


\abstract{It is no secret among deep learning researchers that finding the optimal data augmentation strategy during training can mean the difference between state-of-the-art performance and a run-of-the-mill result. To that end, the AI community has seen many efforts to automate the process of finding the perfect augmentation procedure for any task at hand. Unfortunately, even recent cutting-edge methods bring massive computational overhead, requiring as many as 100 full model trainings to settle on an ideal configuration. We show how to achieve equivalent performance using just 6 trainings with Random Unidimensional Augmentation. Source code is available at \url{https://github.com/fastestimator/RUA/tree/v1.0}.}


\maketitle

\section{Introduction}
\label{sec:Introduction}

Data augmentation is a widely used technique to improve deep-learning model performance. It is sometimes described as a ``freebie'' \cite{yolov4} because it can improve model performance metrics without incurring additional computational costs at inferencing time. Unfortunately, creating a good data augmentation strategy typically requires human expertise and domain knowledge \cite{randAugment}, which is inconvenient during initial development as well as when transferring existing strategies between different tasks. In an effort to overcome these drawbacks, researchers have begun looking for an automated solution to data augmentation.

AutoAugment \cite{autoAugment} and its variants (FastAA \cite{fastAugment} and PBA \cite{pba}) automated the data augmentation process by introducing augmentation parameters which are then jointly optimized alongside the neural network parameters during training. While these methods do offer an automated solution to the problem, they also introduce massive search spaces which in turn significantly increase the time required to train a model. For example, AutoAugment uses Reinforcement Learning (RL) on a search space of size $10^{32}$, which costs thousands of GPU hours to find a solution for a single task. Although later methods such as FastAA and PBA greatly improved the search and reduced computation requirements through data subsampling, they can still be undesirable due to the complexity of implementing joint optimization algorithms.

RandAugment \cite{randAugment} took a different approach by completely removing the policy optimization while achieving better results than prior methods. Unlike its predecessors which rely on applying RL to a search space of size $10^{32}$, RandAugment uses only two global parameters, reducing the search space from $10^{32}$ to $10^2$ so that a grid search can be a simple yet viable solution to the problem. As a result, RL is no longer needed for the policy search, making the method significantly easier to implement and more computationally feasible for practical usage.

Despite the significant complexity and efficiency enhancements made by RandAugment, there is still room for improvement. For example, the default setting of RandAugment uses a 10x10 grid search for the $10^{2}$ search space. While it is technically possible to run any training task 100 times, the computational cost of doing so may still be prohibitive, especially on large-scale datasets.

To reduce costs, a sub-grid is often selected from the 10x10 grid for the actual search. Unfortunately, appropriate sub-grid selection is highly customized to specific problems. This re-introduces a requirement on human expertise and experience, which autonomous methods seek to avoid. For example, for Cifar100 \cite{cifar} the proposed subgrid is $N\in\{1, 2\}, M\in\{2, 6, 10, 14\}$. For ImageNet \cite{imagenet}, a ResNet50 model \cite{resnet} uses the subgrid $N\in\{1, 2, 3\}, M\in\{5, 7, 9, 11, 13, 15\}$, whereas EfficientNet \cite{efficientnet} on the same dataset searches $N\in\{2, 3\}$, and $M\in\{17, 25, 28, 31\}$. Where $M$ represents the global distortion magnitude which controls the intensity of all augmentation operations. $N$ is the number of transformations to be applied in each training step. It is difficult to say what kind of intuition would allow someone to generate such sub-grids for previously unseen problems.

To address these problems, we propose Random Unidimentional Augmentation (RUA): a simpler yet more effective automated data augmentation workflow. The goal of RUA is to achieve the following two objectives:
\begin{enumerate}
   \item Reduce the computational cost required to perform automated search, without sacrificing performance.
   \item Eliminate the need for problem-specific human expertise in the process, enabling a fully automated workflow.
\end{enumerate}

\section{Methods}
\label{sec:methods}

\subsection{Dimensionality reduction: 2D to 1D}
\label{ssec:reduce}

There are 2 global parameters defined in the search space of RandAugment: $M$ and $N$. $M$ represents the global distortion magnitude which controls the intensity of all augmentation operations. $N$ is the number of transformations to be applied in each training step. By default, $M$ and $N$ are both integers ranging from 1 to 10, with 10 giving the maximum augmentation effects.

\begin{figure}[b]
\begin{center}
\includegraphics[width=0.8\linewidth]{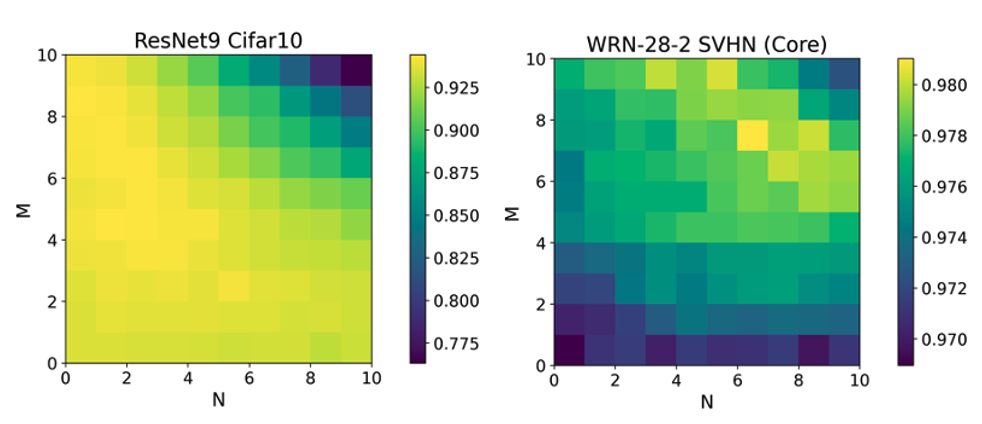}
\end{center}
   \caption{Model accuracy as a function of $M$ and $N$ using RandAugment. Note the (accuracy) gradient as you traverse from the bottom left to the top right of each image.}
\label{fig:grid}
\end{figure}

Although the definitions of $M$ and $N$ are different, the end result of increasing their values is the same: more augmentation. If they could be merged into a single augmentation parameter, then the search space could be reduced by an order of magnitude. To check whether this might be possible, we ran RandAugment on a full 10x10 grid for two classification tasks. We used ResNet9 for Cifar10, and WRN-28-2 \cite{wrn} for SVHN \cite{svhn}. Their test accuracies are shown in Figure \ref{fig:grid}.

The gradients in Figure \ref{fig:grid} show a diagonal trend from the bottom left to the top right. Although the optimal accuracy regions vary between the two problems, the fact that both exhibit an approximately diagonal gradient raises the possibility of traversing the two parameters simultaneously. We confirmed that this pattern is robust to variations in augmentation parameters, as well as across different architectures, tasks, and datasets. These results can be found later on in Figure \ref{fig:newgrid}. We therefore introduce a single parameter $r\in[0,1]$ such that $r=M/M_{max}$ and $r=N/N_{max}$. We then define our augmentation operation parameters directly in terms of $r$, eliminating the need to pick an explicit value for $M_{max}$. This parameterization can be found in Table \ref{tbl:augmentations}. This formulation leaves $N_{max}$ as the single open parameter in the method. While one could simply set $N_{max}=10$ in the footsteps of RandAugment, it can also be set lower while still providing adequate gradient traversal. We defer further discussion of this to Section \ref{ssec:nmax}.

In situations where $r*N_{max}$ is not an integer, we apply $\lfloor r*N_{max} \rfloor$ augmentations, plus a final augmentation which executes with probability equal to the floating point remainder. For example, if $r*N_{max}=3.14$, then 3 augmentations will be guaranteed, and a fourth will execute with 14\% probability.

\subsection{More search with less computation}
\label{ssec:enhance}

\begin{figure}[b]
\begin{center}
\includegraphics[width=0.93\linewidth]{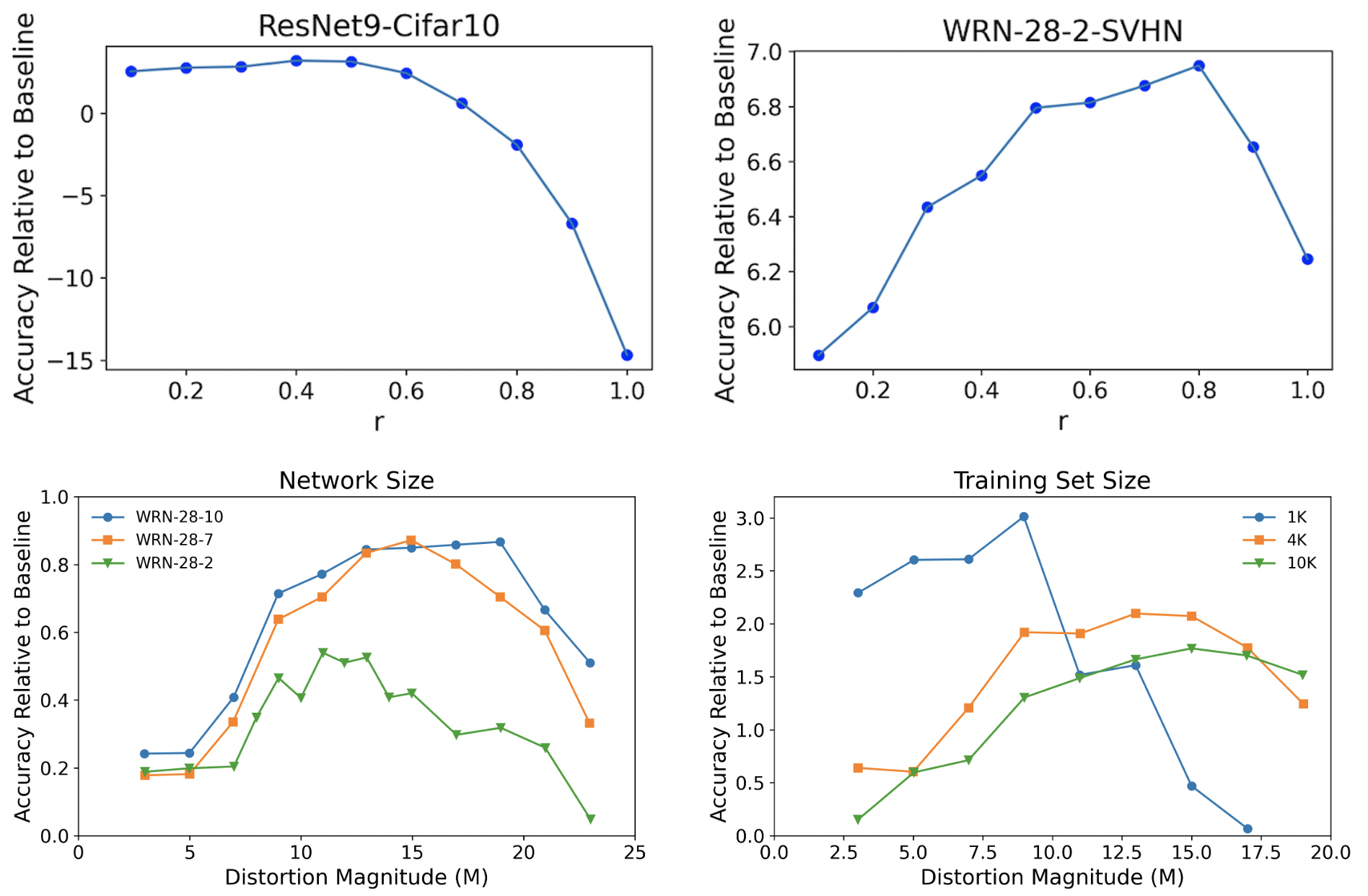}
\end{center}
\caption{RandAugment test accuracy as a function of $r$ ($N_{max}=10$).}
\label{fig:unimode10}
\end{figure}

Another interesting observation one can make from Figure \ref{fig:grid} is that, traversing the diagonals of both Cifar10 and SVHN, accuracy first increases to a maximum and then decreases. In other words, there appears to be unimodality with respect to $r$. If we extract these diagonal terms and plot their relative accuracies against $r$ (Figure \ref{fig:unimode10} top), the unimodal trend becomes more apparent.

The same trend can be observed in the RandAugment paper \cite{randAugment}, reproduced here as Figure \ref{fig:unimode10} bottom. This demonstrates that the unimodal relationship persists across different network and dataset sizes.

In light of this unimodal property, we can leverage algorithms that are more efficient than grid search to explore a larger search space using less computation. One such algorithm is the golden-section search \cite{gss}. Golden-section search is a simple method that is widely used for finding the maximum or minimum of a unimodal function over a given interval. The pseudo code for golden-section search is given in Algorithm \ref{alg:gss}.

\begin{algorithm}
\caption{Golden Section Search (Max-Finding)}
\label{alg:gss}
\begin{algorithmic}
\Require Input function:$f$, range:$[a, b]$, max iterations:$k$
\State $\phi_1, \phi_2 \leftarrow \frac{\sqrt{5} - 1}{2}, \frac{3 - \sqrt{5}}{2}$
\State $h \leftarrow b-a$
\State $c \leftarrow a + \phi_2 * h$
\State $d \leftarrow a + \phi_1 * h$
\State $y_c \leftarrow f(c)$
\State $y_d \leftarrow f(d)$
\For{i from 1 to k}
   \If{$y_c>y_d$}
      \State $b \leftarrow d$
      \State $d \leftarrow c$
      \State $y_d \leftarrow y_c$
      \State $h \leftarrow \phi_1 * h$
      \State $c \leftarrow a + \phi_2 * h$
      \State $y_c \leftarrow f(c)$
   \Else
      \State $a \leftarrow c$
      \State $c \leftarrow d$
      \State $y_c \leftarrow y_d$
      \State $h \leftarrow \phi_1 * h$
      \State $d \leftarrow a + \phi_1 * h$
      \State $y_d \leftarrow f(d)$
   \EndIf
\EndFor
\If{$y_c>y_d$}
   \Return c
\Else
   \Return d
\EndIf
\end{algorithmic}
\end{algorithm}

With golden-section search, every evaluation (after the first) of the search space will reduce the remaining search space by a constant factor of $\approx 0.618$ (inverse golden ratio). As a result, we can search over 90\% of the domain of $r$ using only 6 evaluations. This makes it practical to search over the entire training dataset, without having to resort to subsampling like Fast AA or PBA. Note that this search space reduction does not require any human expertise or intervention, allowing the method to be used as an automated solution for a deep learning task.

\subsection{RUA augmentation parameters}
\label{ssec:align}

After our search space reduction, we are left with one parameter, $r$, which controls the global augmentation intensity. The exact manner of this control is given in Table \ref{tbl:augmentations} (right). A zero value of $r$ means no augmentation, whereas a value of 1 achieves maximum augmentation.

This is a conceptual divergence from RandAugment, as 6 of their 14 transformations are not set up to scale this way. These transformations are marked with a ``*'' in Table \ref{tbl:augmentations} (left). For example, the transformation intensity of \textit{Solarize} and \textit{Posterize} are inversely correlated with $r$. Moreover, \textit{Color}, \textit{Contrast}, \textit{Brightness}, and \textit{Sharpness} are all `shifted' in that they cause no augmentation when $r=0.5$, whereas values closer to 0 or 1 lead to stronger alterations to the input.

In addition to aligning $r$ with augmentation intensity, we also introduce non-deterministic parameter selection into our augmentations. For example, rather than rotating an image exactly $\pm 30$ degrees whenever the \textit{Rotate} operation is applied, we instead draw from a random uniform distribution ($U$) to cover the augmentation space more thoroughly. The maximum intensity of certain augmentations are also increased to keep the expected intensity consistent in spite of the switch to uniform distributions. We justify each of these decisions with an ablation study in Section \ref{ssec:ablation}.

\begin{table*}
\begin{center}
\begin{tabular}{|c|c|c|}
\hline
Augmentations & RandAug (RA) & RUA \\
\hline\hline
Identity & - & - \\
AutoContrast & - & - \\
Equalize & - & - \\
Rotate & degree = $\pm 30r$ & degree = $U(-90r,90r)$ \\
Solarize$^*$ & threshold = $256r$ & threshold = $256 - U(0, 256r)$ \\
Posterize$^*$ & bit shift = $8-4r$ & bit shift = $U(0, 7r)$ \\
Color$^*$ & factor = $1.8r + 0.1$ & factor = $1 + U(-0.9r, 0.9r)$ \\
Contrast$^*$ & factor = $1.8r + 0.1$ & factor = $1 + U(-0.9r, 0.9r)$\\
Brightness$^*$ & factor = $1.8r + 0.1$ & factor = $1 + U(-0.9r, 0.9r)$ \\
Sharpness$^*$ & factor = $1.8r + 0.1$ & factor = $1 + U(-0.9r, 0.9r)$\\
Shear-X & coef = $\pm 0.3r$ & coef = $U(-0.5r, 0.5r)$ \\
Shear-Y & coef = $\pm 0.3r$ & coef = $U(-0.5r, 0.5r)$ \\
Translate-X & coef = $\pm 100r$ & coef = $U(-r, r)*width/3$ \\
Translate-Y & coef = $\pm 100r$ & coef = $U(-r, r)*height/3$ \\
\hline
\end{tabular}
\end{center}
\caption{Augmentations and their associated parameters. Augmentations marked with a ``$*$'' have non-zero impact at $r=0$ under RA, but are zero-aligned under RUA.}
\label{tbl:augmentations}
\end{table*}

\subsection{Selecting a maximum N}
\label{ssec:nmax}

\begin{figure}
\begin{center}
\includegraphics[width=0.67\linewidth]{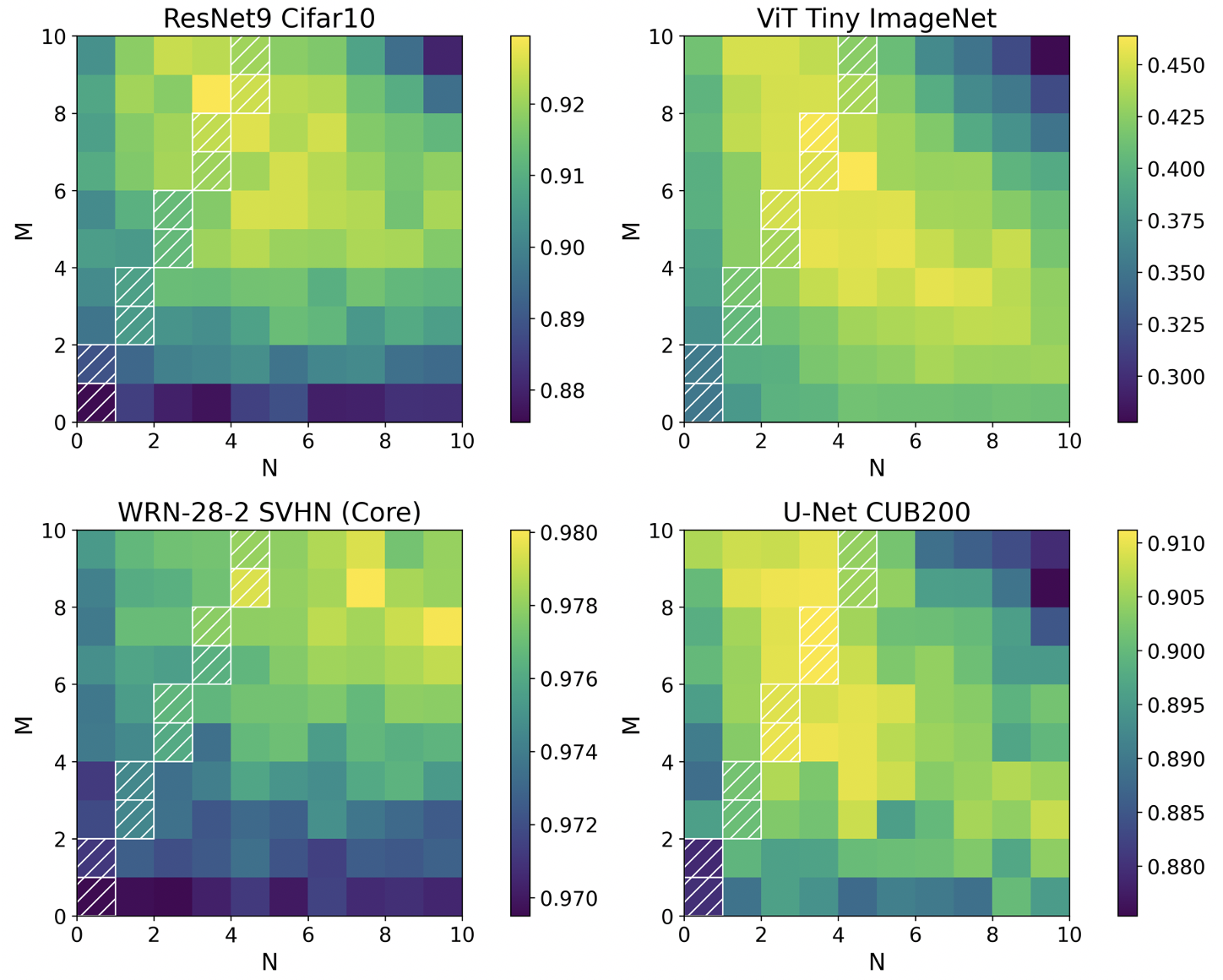}
\end{center}
\caption{Test performance as a function of $M$ and $N$ using RUA augmentation parameters. Note that the (accuracy/dice) gradients as you traverse from the bottom left to the top right of each image are similar to Figure \ref{fig:grid}. Interestingly, this trend exists even in more recent attention-based architectures (ViT\cite{vit} on Tiny ImageNet\cite{tinyImageNet}) and on segmentation tasks (U-Net\cite{unet} on CUB200\cite{cub200}). Cells which are candidates for our RUA search when $N_{max}=5$ are hatched with white.}
\label{fig:newgrid}
\end{figure}

\begin{figure}[H]
\begin{center}
\includegraphics[width=0.45\linewidth]{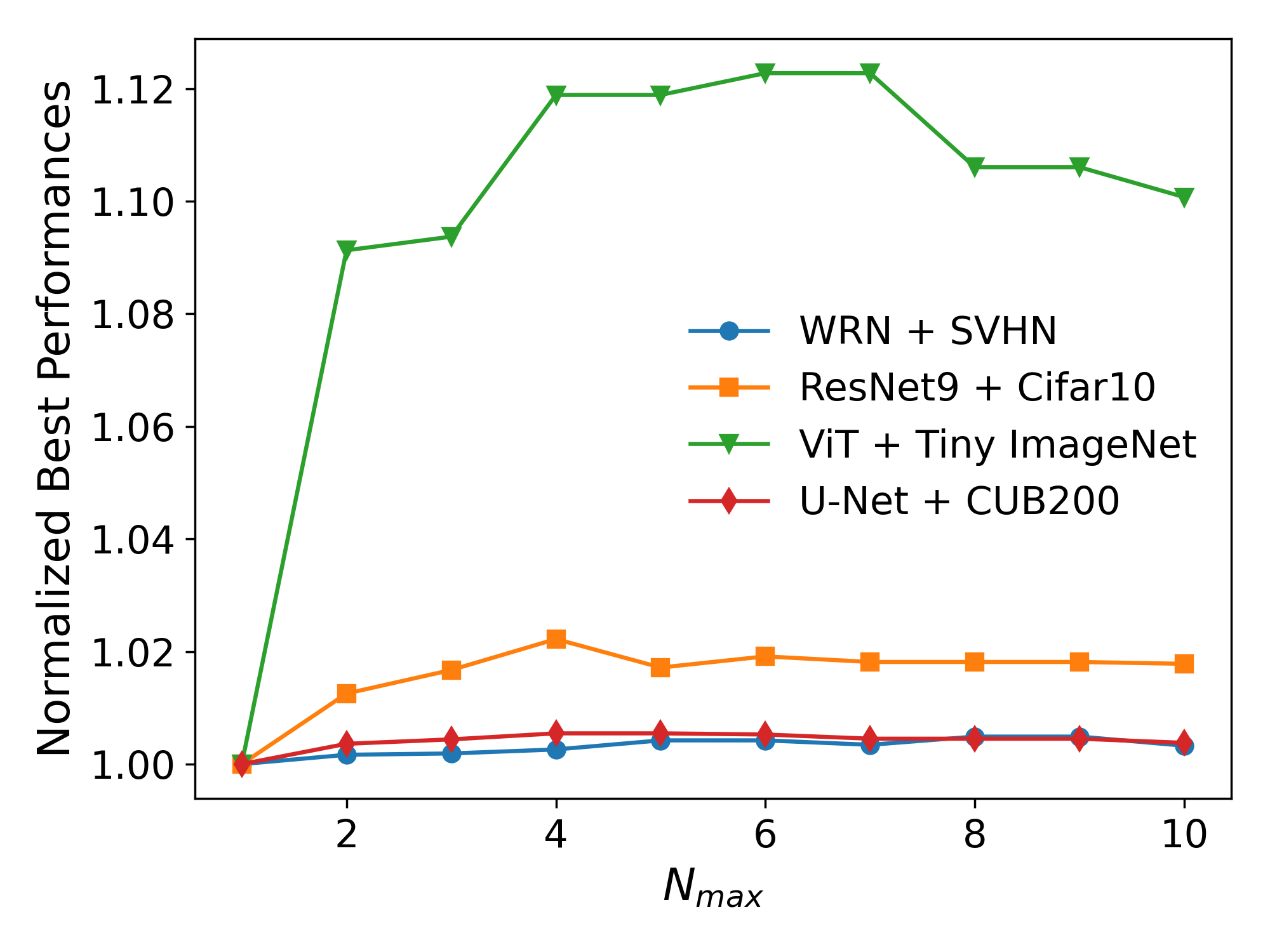}
\end{center}
   \caption{The best performances along a diagonal path as a function of $N_{max}$.}
\label{fig:accnmax}
\end{figure}

One question which must be answered when applying RUA is what value to use as $N_{max}$. While one may be content to use 10, since that was the extent of the RandAugment search space, other numbers may well be equally valid. We ran a second grid search (Figure \ref{fig:newgrid}) using our RUA augmentation parameters to verify that large values of $N_{max}$ may not be necessary in order to achieve a good performance. Based on this search, we examined what outcomes a user would achieve if they ran RUA using different values of $N_{max}$ ranging from 1 to 10. This sensitivity analysis is shown in Figure \ref{fig:accnmax}. In our tests, setting $N_{max} > 5$ does not appear to provide any significant benefit, though small values like 1 or 2 can clearly be harmful, especially for ViT/Tiny ImageNet.

\begin{figure}
\begin{center}
\includegraphics[width=0.5\linewidth]{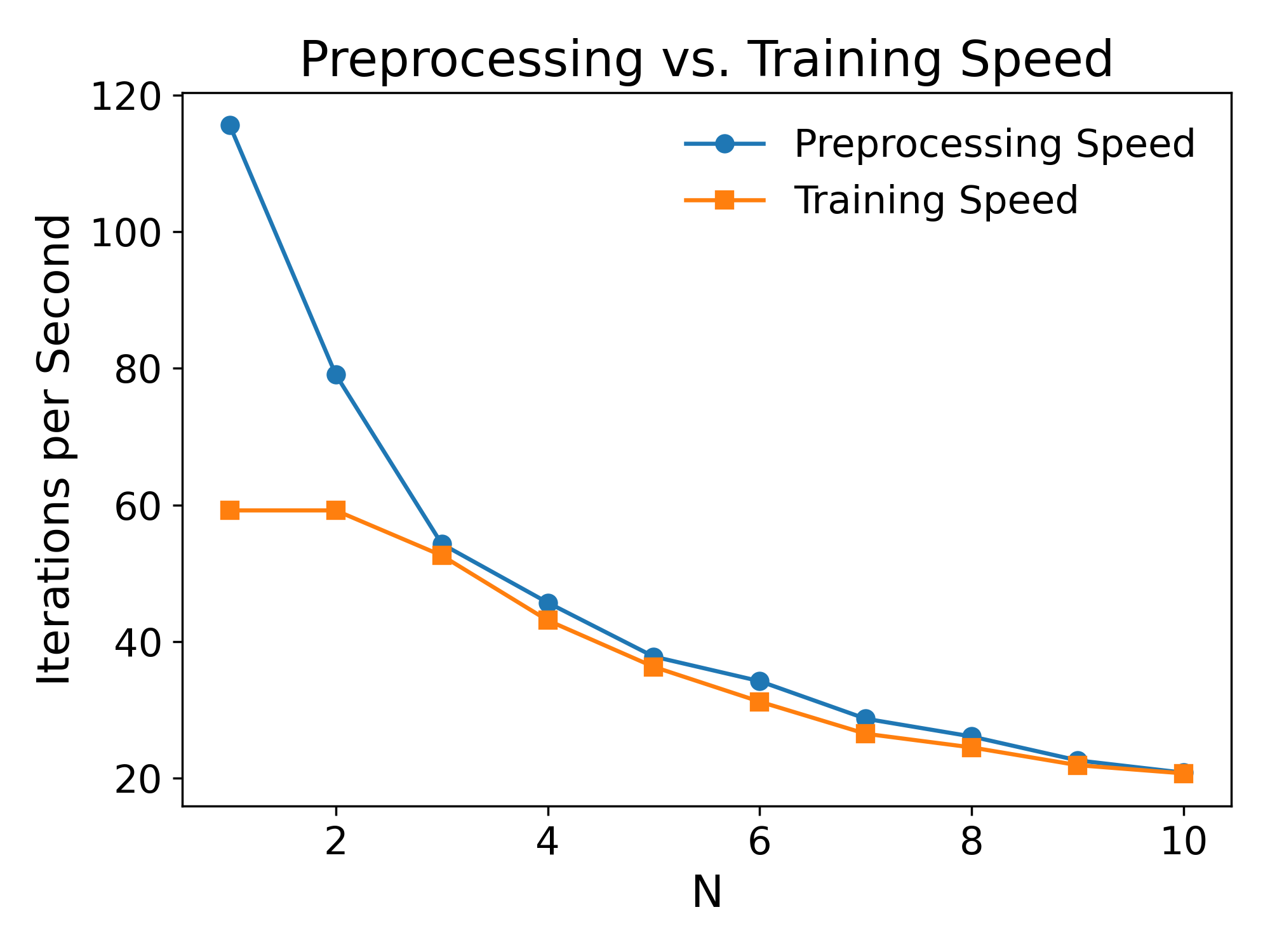}
\end{center}
   \caption{Preprocessing and training speeds as a function of $N$. Measurements were taken on an AWS EC2 P3.2xlarge instance (8 core Intel Xeon CPU, NVIDIA Tesla V100 GPU). Training was conducted using the ResNet9 architecture on Cifar10.}
\label{fig:speed}
\end{figure}

Given that RUA is relatively insensitive to higher values of $N_{max}$, there are pragmatic reasons to choose values smaller than 10. Applying a large number of transformations during training can severely bottleneck the training speed. See Figure \ref{fig:speed} for an example. For our hardware, with any $N\ge 3$ the cpu-based preprocessing became rate limiting, especially once $N\ge 5$. This may be one reason why RandAugment never chose $N>3$ in their sub-grid selections. With these factors in mind, we selected $N_{max}=5$ for our final experiments.

\section{Experiments}
\label{sec:Experiments}

\subsection{RUA performance assessment}
\label{ssec:performance}

In order to perform a direct comparison with previous works, we deploy RUA in the same training setting used by RandAugment on Cifar10, Cifar100, SVHN, and ImageNet. Details regarding the parameters used in each experiment are given in Table \ref{tbl:parameters}. There are a few things worth highlighting about our experimental parameters:

\begin{table}[b]
\begin{center}
\begin{tabular}{|C{0.167\textwidth}||C{0.14\textwidth}|C{0.12\textwidth}|C{0.12\textwidth}|C{0.105\textwidth}|C{0.115\textwidth}|}
\hline
Dataset & CIFAR10 & CIFAR10 & CIFAR100 & SVHN (Core) & ImageNet \\
\hline
Network & PyramidNet-272-200 & Wide-ResNet-28-10 & Wide-ResNet-28-10 & Wide-ResNet-28-2 & ResNet50\\
\hline
Epochs & 900 & 200 & 200 & 200 & 180\\
\hline
Batch Size & 128 & 128 & 128 & 128 & 4096\\
\hline
Image Preprocessing & mean-std-Normalize & mean-std-Normalize & mean-std-Normalize & Divide by 255 & None \\
\hline
Augmentations & [pad-and-crop, horizontal flip, RUA, Cutout] &[pad-and-crop, horizontal flip, RUA, Cutout] & [pad-and-crop, horizontal flip, RUA, Cutout]& [RUA, Cutout] & [random resized crop, horizontal flip, RUA]\\
\hline
Optimizer & SGD & SGD & SGD & SGD & SGD \\
\hline
Weight Decay & \num{1e-4} & \num{5e-4} & \num{5e-4} & \num{5e-4} & \num{1e-4} \\
\hline
Initial LR & 0.1 & 0.1 & 0.1 & 0.1 & 1.6 \\
\hline
LR Schedule & Cosine Decay & Cosine Decay & Cosine Decay & Cosine Decay & $\times0.1$ at epoch 60, 120, and 160\\
\hline
Momentum & 0.9 & 0.9 & 0.9 & 0.9 & 0.9 \\
\hline
$N_{max}$ & 5 & 5 & 5 & 5 & 5\\
\hline
Best $r$ & 0.867 & 0.6 & 0.733 & 0.8 & 0.666 \\
\hline
\end{tabular}
\end{center}
\caption{Experiment parameter details. Note that PyramidNet uses ShakeDrop regularization for consistency with the RandAugment experimental setup.}
\label{tbl:parameters}
\end{table}

\begin{enumerate}
\item In order to be consistent with previous works, we also applied default augmentations before and after applying RUA augmentation on different tasks. For example, pad-and-crop, horizontal flip, and cutout \cite{cutout} are used on the Cifar 10/100 datasets.

\item In Cifar10, RandAugment trained for 1800 epochs whereas the official implementation of PyramidNet \cite{pyramidnet} and ShakeDrop \cite{shakedrop} trained for 300 epochs. We picked 900 epochs as a compromise between different official implementation settings.

\item In every dataset we hold out 5k training samples as evaluation data for selecting the best $r$. After selecting $r$, we put the hold-out set back into the training set and train again. We then record the test performance at the end of that final training.
\end{enumerate}

The final test results of RUA are shown in Table \ref{tbl:results}, where our performance scores are from an average of 10 independent runs. The results of previous methods including the baseline, AA, Fast AA, PBA, and RA are taken from previous work \cite{randAugment}. The best accuracies for each column are highlighted in bold. The search space and the number of iterations reqired by each method is shown in Table \ref{tbl:efficiency}, with the best highlighted in bold.

As demonstrated in Table \ref{tbl:results}, RUA achieved equal or better test scores than previous state-of-the-art methods on 4 out of 5 tasks. For the Cifar10 tasks, we are equivalent to the best prior methods, with one-tailed t-test p-values of 0.0017 and 0.034. For Cifar100 and ImageNet our performance exceeds that of prior methods, with one-tailed t-test p-values of 0.002 and 0.039. On SVHN, despite being outperformed by RandAugment, RUA nonetheless achieved competitive performance on par with AutoAugment.

In addition to the performance, RUA also managed to reduce the search space by an order of magnitude and cut the training iteration requirements of the previous best method by more than 2x as shown in Table \ref{tbl:efficiency}. This proved that the method could optimize the augmentation policy more efficiently than prior methods while achieving equivalent or better results.

\begin{table}
\begin{center}
\begin{tabular}{|c|c|c|c|c|c|}
\hline
\multirow{3}{*}{Methods} & \multicolumn{2}{c|}{CIFAR10} & CIFAR100 & SVHN & ImageNet\\
\cline{2-6}
 & \multirow{2}{*}{PyramidNet} & \multirow{2}{*}{WRN-28-10} & \multirow{2}{*}{WRN-28-10}& \multirow{2}{*}{WRN-28-2} & \multirow{2}{*}{ResNet50}\\
& & &&&\\
\hline
\rule{0pt}{10pt}Baseline & 97.3 & 96.1 & 81.2 & 96.7 & 76.3\\
\hline
\rule{0pt}{10pt}AA & \textbf{98.5} & \textbf{97.4} & 82.9 & 98.0 & 77.6\\
\hline
\rule{0pt}{10pt}Fast AA & 98.3 & 97.3 & 82.7 & - & 77.6\\
\hline
\rule{0pt}{10pt}PBA & \textbf{98.5} & \textbf{97.4} & 83.3 & - & -\\
\hline
\rule{0pt}{10pt}RA & \textbf{98.5} & 97.3 & 83.3 & \textbf{98.3} & 77.6\\
\hline
\rule{0pt}{10pt}RUA & \textbf{98.5} & \textbf{97.4} & \textbf{83.6} & 98.0 & \textbf{77.7}\\
\hline
\end{tabular}
\end{center}
\caption{Experimental results for RUA compared with previous works. We report our average test accuracy over 10 independent runs (as in prior works). Best values in bold.}
\label{tbl:results}
\end{table}

\begin{table}
\begin{center}
\begin{tabular}{|c|C{0.24\linewidth}|C{0.30\linewidth}|}
\hline
Methods & Search Space Order & Search Iterations Required\\
\hline
\rule{0pt}{10pt}AA & $10^{32}$ & $15000$\\
\hline
\rule{0pt}{10pt}Fast AA & $10^{32}$ & $200$\\
\hline
\rule{0pt}{10pt}PBA & $10^{61}$ & $16$\\
\hline
\rule{0pt}{10pt}RA & $10^{2}$ & $100$\\
\hline
\rule{0pt}{10pt}RUA & \textbf{10} & \textbf{6}\\
\hline
\end{tabular}
\end{center}
\caption{The search spaces of various auto-augmentation solutions. Fast AA and PBA search by training on subsampled datasets to improve search speed. Since reduced datasets can be equally applied to any of the above search methods, we directly compare iterations required by each search algorithm rather than dataset/hardware specific metrics. For each method, one iteration involves training the target model to convergence.}
\label{tbl:efficiency}
\end{table}

\subsection{Ablation study}
\label{ssec:ablation}

We conducted an ablation study on the various design decisions outlined in Section \ref{ssec:align}. The results of this study are given in Table \ref{tbl:ablation}. There are several noteworthy takeaways from these comparisons. First, making the ``*'' augmentations from Table \ref{tbl:augmentations} positively correlated with $r$ is always beneficial. This can be seen through pairwise comparisons of rows 1 vs 5, 2 vs 6, 3 vs 7, and 4 vs 8. The second takeaway is that using a random distribution to draw the transformation arguments is always beneficial. This can be seen through pairwise comparisons of rows 1 vs 3, 2 vs 4, 5 vs 7, and 6 vs 8. Finally, increasing the maximum strength of augmentations (for example rotating $\pm 90$ rather than $\pm 30$) is always deleterious on its own (rows 1 vs 2 and 5 vs 6), but advantageous when paired with random sampling (rows 3 vs 4 and 7 vs 8). This is not particularly surprising since larger effects under deterministic sampling will consistently and seriously distort an image, whereas under uniform sampling they permit a larger exploration of the distortion space while still more often sampling less extreme distortions. All told, the best design was to apply all three modifications (row 8).

\begin{table}[b]
\begin{center}
\begin{tabular}{|c|c|c|c||c|}
\cline{2-5}
\multicolumn{1}{c|}{} & Aligned & Random & Expanded & Accuracy \\
\hline
1 & 0 & 0 & 0 & $0.916$ \\
\hline
2 & 0 & 0 & 1 & $0.912$ \\
\hline
3 & 0 & 1 & 0 & $0.917$ \\
\hline
4 & 0 & 1 & 1 & $0.920$ \\
\hline
5 & 1 & 0 & 0 & $0.917$ \\
\hline
6 & 1 & 0 & 1 & $0.915$ \\
\hline
7 & 1 & 1 & 0 & $0.920$ \\
\hline
8 & 1 & 1 & 1 & $0.922$ \\
\hline
\end{tabular}
\end{center}
\caption{An ablation study of the RUA design decisions from Section \ref{ssec:align}. A ResNet9 architecture was trained on Cifar10, with accuracies averaged over 10 independent runs. `Aligned' indicates our modifications to the ``*'' transforms in Table \ref{tbl:augmentations}, `Random' indicates our use of a random uniform distribution, and `Expanded' indicates the use of expanded augmentation parameters. Row 1 is thus analogous to running RandAugment using our dimensionality reduction and golden section search routine, and row 8 is the full RUA method.}
\label{tbl:ablation}
\end{table}

\subsection{Additional findings on out-of-domain generalization}
\label{ssec:additional_finding}

While exploring the proposed RUA transformations on a segmentation problem, we discovered an interesting Out-of-Domain Data (ODD) generalization benefit. ODD generalization problems can arise in machine learning tasks whenever the training data (namely In-Domain-Data: IDD) under-represents the real-world domain. This can result in a model which has high performance within the dataset scope, but poor performance otherwise - even when performing the same task. For example, a segmentation model trained on chest X-Ray images collected from one hospital/region may not generalize well to a different hospital/region. The two datasets might have different distributions for any number of reasons (different quality hardware, different patient demographics, etc.), which may be difficult to account for ahead of time.

Data augmentation can be applied during training to ameliorate this problem to some degree. The hope is that through augmentation the IDD will be expanded to encompass a larger fraction of the data domain, reducing the chance of encountering truly ODD samples. What specific transformations ought to be used to accomplish this, however, remains an open question. One popular school of thought is to select transformations based on domain availability \cite{realistic_aug}. In other words, only select transformations which result in outputs which are `realistic' in the sense that you would expect to find them in the real world. For example, if solarization never occurs in clinically acquired chest X-Ray images, then applying solarization during training would not be recommended for chest X-Ray applications. While domain availability is an intuitively reasonable heuristic, it is potentially at odds with RUA, which proposes searching over a standardized set of augmentations regardless of the target task.

To test whether domain availability is a valuable/necessary heuristic, we designed an experiment using two different datasets: the Montgomery dataset \cite{montgomery} and the NIH dataset\cite{nih_xray}. Both datasets contain Chest X-Ray images with pixel-level mask annotations for lung cavities. To test IDD performance we trained segmentation models on the Montgomery dataset using traditional train/test dataset splitting. To test ODD generalization, we then inferenced the models on the NIH dataset (which is larger and more heterogeneous).

We experimented with 6 different augmentation strategies, with 4 strategies proposed by different prior works for domain-specific applications on Chest X-Ray data. The augmentation baselines include:

\begin{itemize}
   \item \textbf{No Aug} - No data augmentation is applied
   \item \textbf{Method 1 (M1)} - Chest X-Ray transformations as proposed in \cite{aug1}
   \item \textbf{Method 2 (M2)} - Chest X-Ray transformations as proposed in \cite{aug2}
   \item \textbf{Method 3 (M3)} - Chest X-Ray transformations as proposed in \cite{aug3}
   \item \textbf{Method 4 (M4)} - Chest X-Ray transformations as proposed in \cite{aug4}
   \item \textbf{RUA} - Our proposed transformations, applied at their maximum strength
\end{itemize}

\begin{figure}
\begin{center}
\includegraphics[width=0.8\linewidth]{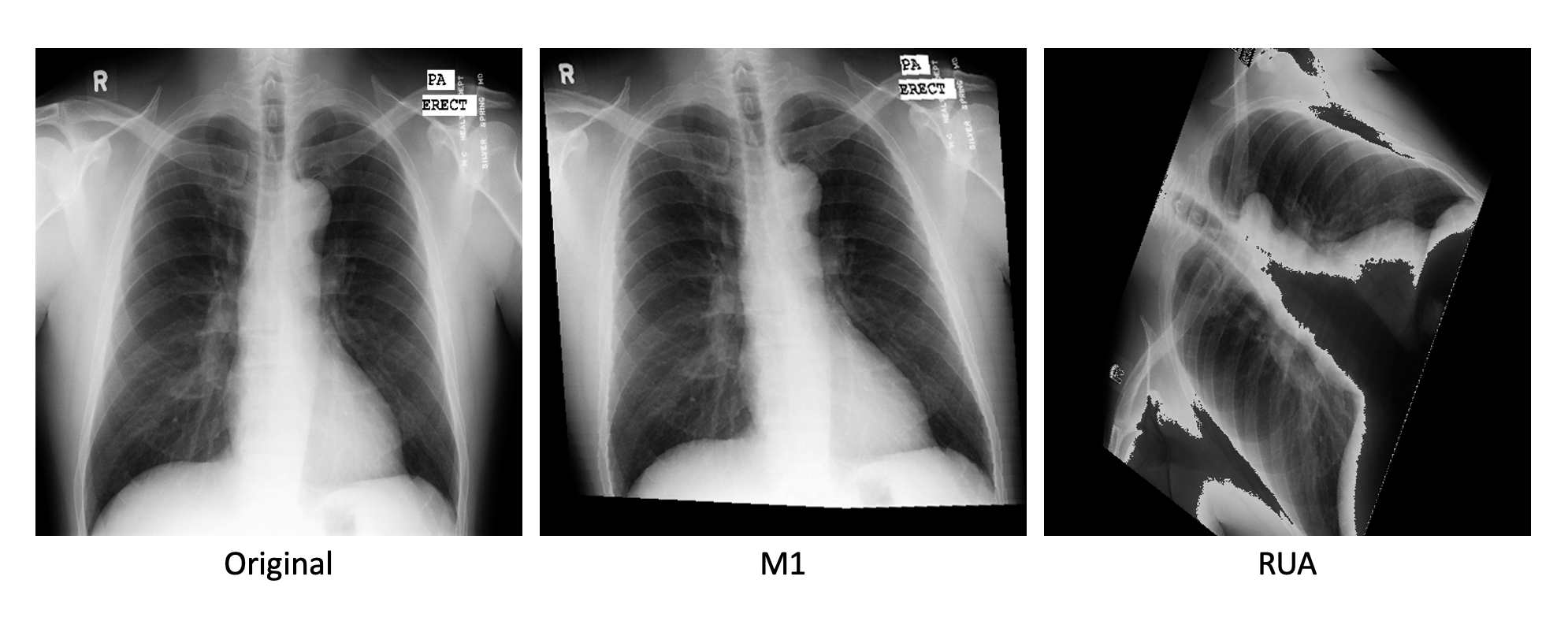}
\end{center}
\caption{A sample image under different augmentation strategies.}
\label{fig:samples}
\end{figure}

While all of the baseline methods adhere to the domain availability heuristic (being comprised of mostly minor rotates, flips, stretching, etc.), our approach results in heavily distorted images that would never be found in a real clinical setting. A representative example can be found in Figure \ref{fig:samples}.

For these experiments we used a U-Net architecture trained via a pixel-level cross entropy loss. The training ran for 480 epochs with 5 checkpoints saved evenly throughout. For each augmentation strategy, all 5 checkpoints were evaluated on the NIH dataset and the checkpoint with the highest average dice score was selected for comparison against other strategies. We report both the IDD and ODD dice scores for each strategy in Table \ref{tbl:best}, along with ODD dice score histogram counts. The sample-wise breakdown of dice scores for each strategy is also visualized in Figure \ref{fig:odd}. Detailed evaluation results for all checkpoints and strategies are provided in Appendix Table \ref{tbl:96epochs}, \ref{tbl:128epochs}, \ref{tbl:288epochs}, \ref{tbl:384epochs}, \ref{tbl:480epochs}.

\begin{table}[h]
\begin{center}
\begin{tabular}{|c|c|c|c|c|c|c|}
\cline{2-7}
\multicolumn{1}{c|}{} & No Aug & M1 & M2 & M3 & M4 & RUA \\
\hline
IDD Avg Dice & $98\pm2.0$ & $98\pm1.5$ & $98\pm1.9$ & $98\pm1.6$ & $98\pm1.7$ & $98\pm0.8$ \\
\hline
ODD Avg Dice & $85\pm12$ & $88\pm11$ & $86\pm14$ & $86\pm13$ & $84\pm17$ & $94\pm5.0$ \\
\hline
\hline
ODD Dice Bins & No Aug & M1 & M2 & M3 & M4 & RUA \\
\hline
0.0 - 0.2 & 18 & 15 & 47 & 30 & 83 & 0 \\
\hline
0.2 - 0.4 & 43 & 27 & 62 & 44 & 88 & 2 \\
\hline
0.4 - 0.6 & 141 & 86 & 155 & 161 & 212 & 10 \\
\hline
0.6 - 0.8 & 923 & 642 & 656 & 757 & 743 & 110 \\
\hline
0.8 - 1.0 & 3841 & 4196 & 4046 & 3974 & 3840 & 4844 \\
\hline
\end{tabular}
\end{center}
\caption{Best Dice scores and distributions for various augmentation strategies.}
\label{tbl:best}
\end{table}

\begin{figure}
\begin{center}
\includegraphics[width=0.8\linewidth]{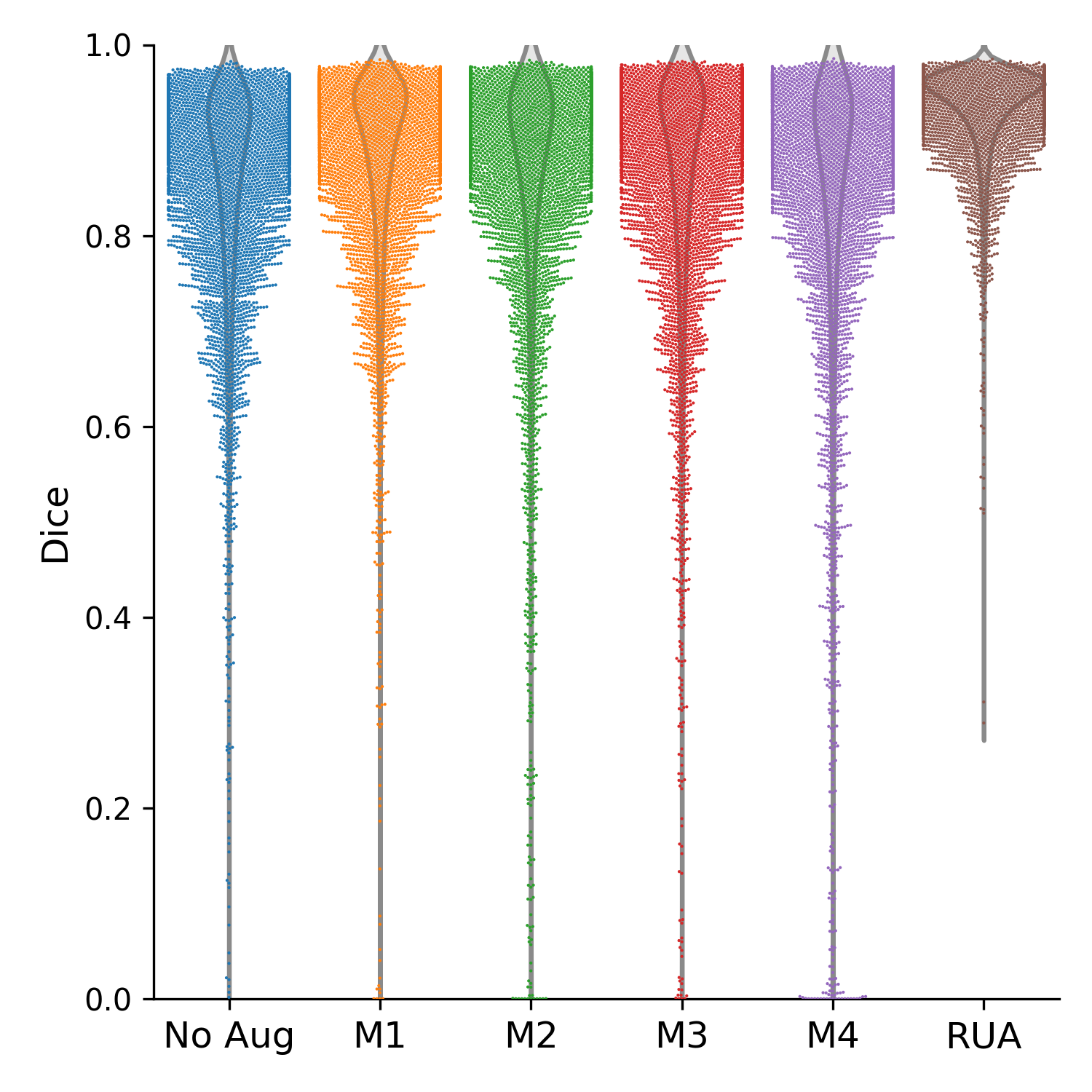}
\end{center}
\caption{A swarm plot (points corresponding to individual data sample performance) overlaid by a violin plot (showing distributional performance measures) on ODD data for different augmentation strategies.}
\label{fig:odd}
\end{figure}

As Table \ref{tbl:best} demonstrates, every strategy suffers reduced ODD performance compared to their nearly perfect IDD performance. The size of this gap, however, varies significantly between different baselines. In particular, RUA's augmentation strategy greatly outperforms all of the `domain availability' augmentations in spite of its less realistic outputs.

The qualitative nature of the difference between RUA and the other methods can be seen in Figure \ref{fig:odd}. The swarm plot tail (outlier cases where the model mostly failed) is significantly smaller for RUA than any of the other strategies (especially for dice scores < 0.7). As the overlaid violin plot indicates, RUA also shifts the mean performance up significantly, with predictions centered around a dice score of 94 and a much smaller standard deviation than the other methods. These improvements are observed in every checkpoint, as demonstrated in Appendix \ref{secA1}, Table \ref{tbl:96epochs}, \ref{tbl:128epochs}, \ref{tbl:288epochs}, \ref{tbl:384epochs}, \ref{tbl:480epochs}.

This finding suggests a new perspective for viewing data augmentation. The fact that the network can learn from heavily distorted (even unrealistic) images in order to solve real-world ODD generalization problems may imply that data augmentation offers a more generalized benefit than simply teaching the model to be invariant to whatever specific transformation has been applied. There is presumably some bound/limit on how much unrealistic data augmentation can be added before causing more harm than good, which is something we plan to investigate further in future work.

\section{Conclusion}
\label{sec:conclusion}

In this work, we proposed Random Unidimensional Augmentation (RUA), an automated augmentation method providing several benefits relative to previous state-of-the-art algorithms. Our search space is one order of magnitude smaller than prior works, our transformations are more effective, and we leverage more efficient search algorithms. As a result of these improvements, RUA achieves equivalent results while requiring significantly less computation. We experimentally demonstrated RUAs strength on the same tasks used by previous works across various network architectures and datasets. Unlike previous methods, RUA does not rely on any problem-specific human expertise, making the method truly automated and thus fit for use in conjunction with larger autoML pipelines. We also showed that automated data augmentation need not necessarily produce realistic data samples to have a positive impact on a network's generalization performance.

\backmatter

\section*{Conflict of Interest}

The authors declare that they have no conflict of interest.

\section*{Data Availability Statement}
The datasets used in this study are publicly available for download in their corresponding websites. The source code to reproduce this work has been open-sourced and can be found at \url{https://github.com/fastestimator/RUA/tree/v1.0}.

\begin{appendices}

\section{Unabridged Segmentation Results}\label{secA1}

\begin{table}[h]
\begin{center}
\begin{tabular}{|c|c|c|c|c|c|c|}
\cline{2-7}
\multicolumn{1}{c|}{} & No Aug & M1 & M2 & M3 & M4 & RUA \\
\hline
IDD Avg Dice & $98\pm2.0$ & $98\pm1.4$ & $98\pm1.6$ & $98\pm1.6$ & $98\pm2.0$ & $98\pm0.9$ \\
\hline
ODD Avg Dice & $84\pm15$ & $87\pm11$ & $85\pm17$ & $86\pm13$ & $83\pm16$ & $91\pm7.0$ \\
\hline
\hline
ODD Dice Bins & No Aug & M1 & M2 & M3 & M4 & RUA \\
\hline
0 - 0.2 & 50 & 21 & 79 & 30 & 65 & 0 \\
\hline
0.2 - 0.4 & 73 & 24 & 99 & 44 & 92 & 0 \\
\hline
0.4 - 0.6 & 219 & 110 & 198 & 161 & 234 & 20 \\
\hline
0.6 - 0.8 & 858 & 661 & 663 & 757 & 926 & 370 \\
\hline
0.8 - 1 & 3766 & 4150 & 3927 & 3974 & 3649 & 4576 \\
\hline
\end{tabular}
\end{center}
\caption{Dice scores for models trained for 96 epochs using various augmentation procedures when tested on out of domain data.}
\label{tbl:96epochs}
\end{table}

\begin{table}[h]
\begin{center}
\begin{tabular}{|c|c|c|c|c|c|c|}
\cline{2-7}
\multicolumn{1}{c|}{} & No Aug & M1 & M2 & M3 & M4 & RUA \\
\hline
IDD Avg Dice & $98\pm2.0$ & $98\pm1.6$ & $98\pm1.9$ & $98\pm2.0$ & $98\pm1.7$ & $98\pm1.2$ \\
\hline
ODD Avg Dice & $85\pm13$ & $86\pm13$ & $86\pm14$ & $86\pm14$ & $84\pm17$ & $89\pm8.7$ \\
\hline
\hline
ODD Dice Bins & No Aug & M1 & M2 & M3 & M4 & RUA \\
\hline
0 - 0.2 & 23 & 34 & 47 & 36 & 83 & 0 \\
\hline
0.2 - 0.4 & 40 & 51 & 62 & 63 & 88 & 3 \\
\hline
0.4 - 0.6 & 169 & 136 & 155 & 160 & 212 & 64 \\
\hline
0.6 - 0.8 & 921 & 711 & 656 & 770 & 743 & 615 \\
\hline
0.8 - 1 & 3813 & 4034 & 4046 & 3937 & 3840 & 4284 \\
\hline
\end{tabular}
\end{center}
\caption{Dice scores for models trained for 182 epochs using various augmentation procedures when tested on out of domain data.}
\label{tbl:128epochs}
\end{table}

\begin{table}[h!]
\begin{center}
\begin{tabular}{|c|c|c|c|c|c|c|}
\cline{2-7}
\multicolumn{1}{c|}{} & No Aug & M1 & M2 & M3 & M4 & RUA \\
\hline
IDD Avg Dice & $98\pm2.1$ & $98\pm1.9$ & $98\pm2.0$& $98\pm2.3$ & $98\pm1.7$ & $98\pm1.1$ \\
\hline
ODD Avg Dice & $85\pm13$ & $86\pm12$ & $85\pm14$ & $86\pm13$ & $83\pm18$ & $93\pm5.7$ \\
\hline
Dice Std & 13.0 & 12.2 & 14.3 & 13.0 & 17.7 & 5.7 \\
\hline
\hline
ODD Dice Bins & No Aug & M1 & M2 & M3 & M4 & RUA \\
\hline
0 - 0.2 & 30 & 21 & 55 & 28 & 103 & 0 \\
\hline
0.2 - 0.4 & 40 & 37 & 62 & 46 & 97 & 0 \\
\hline
0.4 - 0.6 & 181 & 121 & 153 & 153 & 235 & 10 \\
\hline
0.6 - 0.8 & 918 & 792 & 754 & 774 & 762 & 199 \\
\hline
0.8 - 1 & 3797 & 3995 & 3942 & 3965 & 3769 & 4757 \\
\hline
\end{tabular}
\end{center}
\caption{Dice scores for models trained for 288 epochs using various augmentation procedures when tested on out of domain data.}
\label{tbl:288epochs}
\end{table}

\begin{table}[h]
\begin{center}
\begin{tabular}{|c|c|c|c|c|c|c|}
\cline{2-7}
\multicolumn{1}{c|}{} & No Aug & M1 & M2 & M3 & M4 & RUA \\
\hline
IDD Avg Dice & $98\pm2.1$ & $98\pm1.5$ & $98\pm2.1$ & $98\pm2.1$ & $98\pm1.9$ & $98\pm0.8$ \\
\hline
ODD Avg Dice & $85\pm13$ & $88\pm11$ & $85\pm15$ & $86\pm13$ & $83\pm18$ & $94\pm5.0$ \\
\hline
\hline
ODD Dice Bins & No Aug & M1 & M2 & M3 & M4 & RUA \\
\hline
0 - 0.2 & 30 & 15 & 65 & 28 & 109 & 0 \\
\hline
0.2 - 0.4 & 48 & 27 & 61 & 48 & 101 & 2 \\
\hline
0.4 - 0.6 & 176 & 86 & 176 & 143 & 231 & 10 \\
\hline
0.6 - 0.8 & 929 & 642 & 749 & 738 & 811 & 110 \\
\hline
0.8 - 1 & 3783 & 4196 & 3915 & 4009 & 3714 & 4844 \\
\hline
\end{tabular}
\end{center}
\caption{Dice scores for models trained for 384 epochs using various augmentation procedures when tested on out of domain data.}
\label{tbl:384epochs}
\end{table}

\begin{table}[h]
\begin{center}
\begin{tabular}{|c|c|c|c|c|c|c|}
\cline{2-7}
\multicolumn{1}{c|}{} & No Aug & M1 & M2 & M3 & M4 & RUA \\
\hline
IDD Avg Dice & $98\pm2.0$ & $98\pm2.0$ & $98\pm2.0$ & $98\pm2.2$ & $98\pm1.7$ & $98\pm0.6$ \\
\hline
ODD Avg Dice & $85\pm12$ & $87\pm13$ & $84\pm15$ & $86\pm14$ & $83\pm18$ & $92\pm6.4$ \\
\hline
\hline
ODD Dice Bins & No Aug & M1 & M2 & M3 & M4 & RUA \\
\hline
0 - 0.2 & 18 & 29 & 72 & 32 & 119 & 0 \\
\hline
0.2 - 0.4 & 43 & 40 & 67 & 50 & 95 & 0 \\
\hline
0.4 - 0.6 & 141 & 121 & 183 & 169 & 236 & 18 \\
\hline
0.6 - 0.8 & 923 & 695 & 744 & 752 & 831 & 264 \\
\hline
0.8 - 1 & 3841 & 4081 & 3900 & 3963 & 3685 & 4684 \\
\hline
\end{tabular}
\end{center}
\caption{Dice scores for models trained for 480 epochs using various augmentation procedures when tested on out of domain data.}
\label{tbl:480epochs}
\end{table}

\end{appendices}

\bibliography{rua}



\end{document}